\documentclass[letterpaper, 10 pt, conference]{ieeeconf}  

\IEEEoverridecommandlockouts                              

\usepackage{bm}

\usepackage[font=small,labelfont=bf]{caption}
\usepackage{booktabs}
\usepackage{multirow}
\usepackage{colortbl}
\usepackage{amssymb}
\usepackage{amsmath}
\usepackage{csquotes}
\usepackage[breaklinks,colorlinks=true,linkcolor=red, citecolor=blue]{hyperref}%
\usepackage[capitalise]{cleveref}
\crefname{table}{Tab.}{Tabs.}
\Crefname{table}{Table}{Tables}    
\crefname{section}{Sec.}{Secs.}
\usepackage[hyperref=true, backref=false, sorting=none]{biblatex}
\usepackage[dvipsnames]{xcolor}

\colorlet{colorFst}{Green!40}       
\colorlet{colorSnd}{SpringGreen!50} 
\colorlet{colorTrd}{Yellow!40}      
\colorlet{colorLow}{darkgray!30}    
\newcommand{\fs}{\cellcolor{colorFst}\bf}   
\newcommand{\nd}{\cellcolor{colorSnd}}      
\newcommand{\rd}{\cellcolor{colorTrd}}      

\newcommand{\RThree}[0]{\mathbb{R}^{3}}
\newcommand{\RFour}[0]{\mathbb{R}^{4}}

\usepackage{arydshln}

\usepackage[colorinlistoftodos]{todonotes}

\makeatletter
\def\footnoterule{\kern-3\p@
  \hrule \@width 2in \kern 2.6\p@} 
\makeatother

\usepackage[Export]{adjustbox} 

\usepackage{bbding}
\usepackage{makecell}

\usepackage{subcaption}

\title{\LARGE \bf
High-Fidelity SLAM Using Gaussian Splatting with Rendering-Guided Densification and Regularized Optimization
}

\author{Shuo Sun$^{1}$, Malcolm Mielle$^{2}$, Achim J. Lilienthal$^{1,3}$, and Martin Magnusson$^{1}$
\thanks{*This work has received funding from the European Union’s Horizon 2020 research and innovation programme under grant agreement No 101017274 (DARKO).}
\thanks{$^{1}$AASS research center, \"{O}rebro University, Sweden.
    {\tt\small 
        \{shuo.sun, achim.lilienthal, martin.magnusson\}@oru.se
    }
}%
\thanks{$^{2}$Independent researcher.
        {\tt\small {malcolm.mielle@protonmail.com}}
}%
\thanks{$^{3}$Technical University of Munich, Chair: Perception for Intelligent Systems.
    {\tt\small 
        achim.j.lilienthal@tum.de
    }
}%
}

\AtBeginBibliography{\scriptsize}
\begin{document}

\maketitle
\thispagestyle{empty}
\pagestyle{empty}

\begin{abstract}
We propose a dense RGBD SLAM system based on 3D Gaussian Splatting that provides metrically accurate pose tracking and visually realistic reconstruction.
To this end, we first propose a Gaussian densification strategy based on the rendering loss to map unobserved areas and refine reobserved areas.
Second, we introduce extra regularization parameters to alleviate the ``forgetting" problem during contiunous mapping, where parameters tend to overfit the latest frame and result in decreasing rendering quality for previous frames.
Both mapping and tracking are performed with Gaussian parameters by minimizing re-rendering loss in a differentiable way.
Compared to recent neural and concurrently developed Gaussian splatting RGBD SLAM baselines, our method achieves state-of-the-art results on the synthetic dataset \texttt{Replica} and competitive results on the real-world dataset \texttt{TUM}.
The code is released on
\href{https://github.com/ljjTYJR/HF-SLAM}{\color{blue}{\textbf{$\texttt{https://github.com/ljjTYJR/HF-SLAM}$}}}.

\end{abstract}

\begin{figure*}
    \centering
    \includegraphics[width=0.9\textwidth]{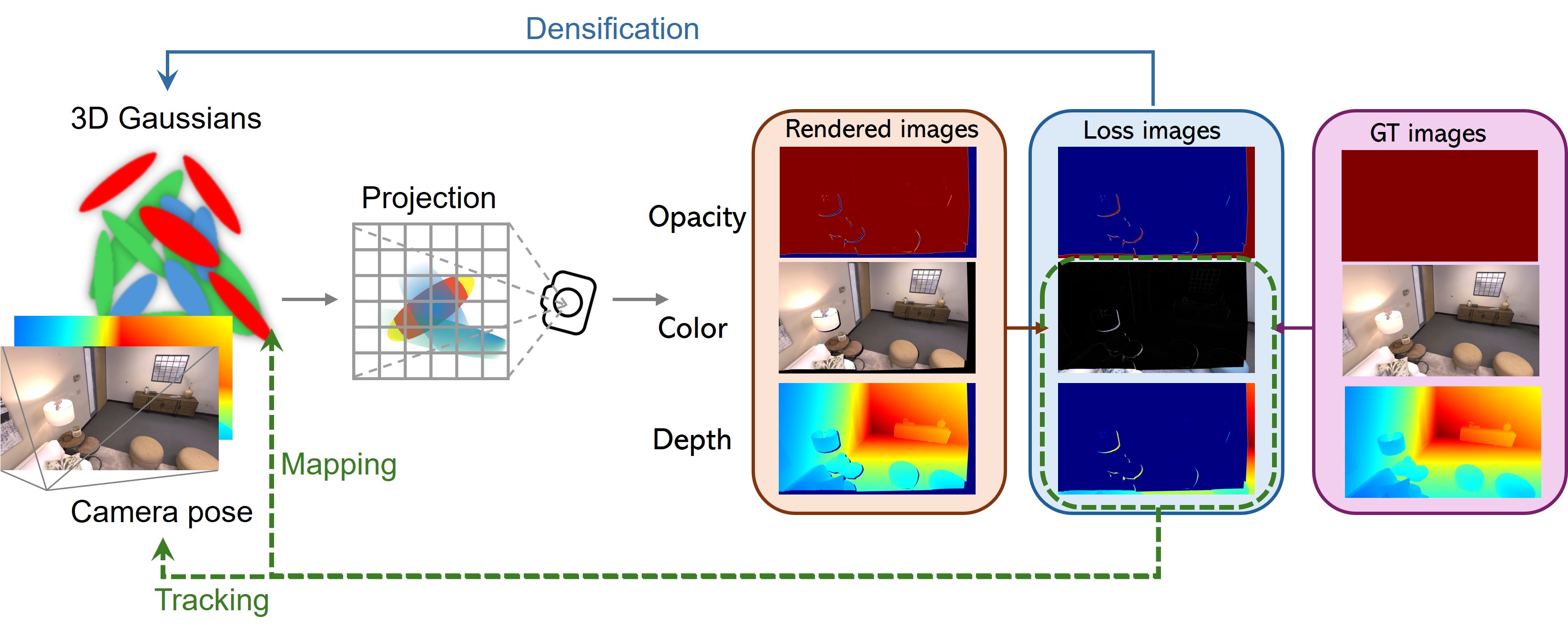}
    \caption{
        \textbf{Overview of the method.}
        Our method takes RGBD frames as inputs.
        During mapping, when given a posed RGBD frame, we first render the opacity image, color image and depth image.
        Then we compare them with the ground truth to densify the existed map.
        During tracking, we minimize the color and depth re-rendering loss to optimize the camera pose.
    }
    \label{fig:enter-label}
\end{figure*}
\section{Introduction}
Dense visual SLAM with RGBD inputs is essential for many downstream tasks in mobile robots, AR/VR and robot manipulation.
Traditional SLAM systems have focused mainly on camera tracking~\cite{mur2015orb} and geometric surface reconstruction~\cite{newcombe2011kinectfusion}.
However, surface appearance reconstruction, which contains rich information for scene understanding, is often lacking in traditional SLAM systems.
Motivated by the success of \emph{NeRF}~\cite{yen2021inerf} on novel view synthesis, recent RGBD SLAM methods use volumetric structures~\cite{zhu2022nice,johari2023eslam,wang2023co,sucar2021imap} or points~\cite{sandstrom2023point} as the map representation.
Combined with neural networks as decoders, these methods can achieve both appearance and geometry reconstruction.
However, limited by the computationally expensive sampling along rays in the optimization of NeRF, the above mentioned methods cannot render full-resolution images with high quality (commonly only 0.5\% pixels are used for mapping~\cite{johari2023eslam}, leading to low quality image rendering; see \cref{fig:rendering_results_on_replica} and \cref{tab:reconstruction_performance_replica}).

The recent work \emph{3D Gaussian Splatting}~(3DGS)~\cite{kerbl20233dgaussian} 
provides a more efficient way for novel view synthesis.
Instead of expensive sampling along the ray, 3DGS relies on rasterization for rendering, which accelerates the mapping process so as to handle full-resolution images.
The original 3DGS needs prior camera poses, which are often estimated by \emph{Structure-from-Motion}~\cite{schonberger2016structure}.
In this work, we extend 3DGS to the case of online tracking and mapping, removing the need for prior camera poses.
Our method processes sequential RGBD frame inputs, simultaneously optimizing Gaussian parameters and estimating camera poses.
Though there are some concurrent works combining 3DGS with SLAM~\cite{keetha2023splatam,yan2023gs}, we show our method achieves better or competitive rendering and tracking results compared to recent baselines.


Our first contribution is the introduction of a novel densification strategy based on rendering. We directly densify the map utilizing rendering loss, enabling us to effectively map unobserved areas and enhance the rendering quality of reobserved regions.

Our second contribution is regularized optimization during mapping.
Continuous mapping with 3DGS suffers from the ``forgetting" problem, which means that Gaussian parameters tend to overfit to the latest frame and lead to decreased reconstruction quality for previous frames (see \cref{fig:forgetting-demonstration}).
To alleviate the forgetting problem, we introduce extra parameters to supervise the learning process, and our experiments show that our method can preserve the rendering quality of previously visited areas.

We demonstrate state-of-the-art reconstruction quality and tracking accuracy on $\texttt{Replica}$ dataset and competitive results on the real-world dataset $\texttt{TUM-RGBD}$.
We show the relative effectiveness of the two main contributions in the ablation study. 
We also find that continuous mapping can reconstruct high-quality images even if when faced with motion blur, which may inspire work on 3DGS reconstruction with blurry inputs.

\section{Related Work}
\subsection{Visual SLAM}
Conventional visual SLAM methods are divided into direct and indirect methods.
Indirect methods (e.g., \emph{ORB-SLAM}~\cite{mur2015orb}) first detect points of interest and attach feature descriptors.
Tracking is conducted by feature matching and minimizing the re-projection error across frames.
Direct methods (e.g., \emph{DSO}~\cite{engel2017direct}) consume raw pixels and construct photometric loss for tracking.
These conventional visual SLAM methods focus more on tracking instead of scene reconstruction.
Since \emph{NeRF}~\cite{mildenhall2021nerf} has shown powerful scene reconstruction abilities, recent work applies neural representation in  SLAM to produce more complete and photorealistic scene reconstruction.
For example, \emph{iMAP}~\cite{sucar2021imap} uses a neural network to represent the scene; \emph{NICE-SLAM}~\cite{zhu2022nice} and later work~\cite{johari2023eslam,wang2023co} combine neural networks and voxel grids to improve reconstruction quality.
Thanks to the generalization capabilities of neural networks, these neural representations have more powerful abilities to fill holes and construct more complete scenes compared to explicit representations.
Although neural representations achieve good results in geometric scene reconstruction, their performance in recovering high-fidelity scenes is not satisfactory enough.
In this work, we use parameterized Gaussians~\cite{kerbl20233dgaussian} as the map representation and achieve a more photorealistic reconstruction.
Concurrent with our work, some SLAM systems are also based on Gaussian Splatting: \emph{MonoGS}~\cite{matsuki2023gaussian} expands the map by random sampling; \emph{Gaussian-SLAM}~\cite{yugay2023gaussian} divides the scene into many submaps; \emph{Photo-SLAM}~\cite{huang2023photo} conducts tracking by minimizing the reprojection error; \emph{SplaTAM}~\cite{keetha2023splatam} and \emph{GS-SLAM}~\cite{yan2023gs} densify the map only considering the unobserved regions (more information can be found in \cite{chen2024survey}).
Our method expands the map by considering filling holes in unobserved areas and refining reobserved areas, which can effectively improve the reconstruction quality.
In addition, our regularized optimization can effectively alleviate the forgetting problem during mapping and preserve details of previously visited areas.

\subsection{Photo-realistic reconstruction}
\emph{NeRF}~\cite{mildenhall2021nerf} opens a convenient way for photo-realistic reconstruction from only 2D images.
Based on neural networks and volumetric differentiable rendering, NeRF can render novel photorealistic views that were not observed in the inputs.
Many follow-up works improve on the original NeRF by accelerating training speed~\cite{muller2022instant}, improving rendering quality~\cite{barron2023zip}, {etc.}
Some works~\cite{fridovich2022plenoxels} also reveal that neural networks are not necessary for novel view synthesis.
As opposed to Gaussian splatting~\cite{kerbl20233dgaussian}, however, all the above-mentioned methods adopt expensive ray-marching when calculating one pixel, resulting in low rendering speed (lower than 15 fps).
Gaussian Splatting uses parameterized Gaussians (as introduced in \cref{sec:gaussian_splatting_rendering}) as rendering primitives.
Instead of expensive ray-marching, Gaussian Splatting renders the image by tile-based rasterization, which can achieve much faster rendering speed (around 100 fps).
The original Gaussian Splatting needs known camera poses (commonly achieved by \emph{SfM}~\cite{schonberger2016structure}) and optimizes Gaussian parameters offline.
In this work, we consider sequential RGBD frame inputs and optimize parameters online, which releases the requirement of known camera poses and brings the possibility of online applications such as exploration.


\section{Preliminary: Gaussian Splatting Rendering} \label{sec:gaussian_splatting_rendering}


We use 3D Gaussians as the map representation primitives for the color and depth image rendering.
As described in \cite{kerbl20233dgaussian}, each Gaussian consists of parameters 
\begin{equation}
    \mathbf{\Theta}_i \mathrel{\mathop:}= \{\bm{\mu}_{i}, \bm{s}_{i}, \bm{r}_{i}, \bm{c}_{i}, o_i \}, \label{eq:gaussian_param}
\end{equation}
with Gaussian means $\bm{\mu}_{i} \in \RThree$, scales $\bm{s_{i}} \in \RThree$, rotation quaternion $\bm{r}_{i} \in \RFour$, RGB color $\bm{c}_{i} \in \RThree$, and opacity $o_{i} \in \mathbb{R}$. All of these parameters are optimizable during training.

When rendering the image at some camera pose $T$, the 3D Gaussians in the field of view are projected onto the image plane as 2D Gaussians with mean $\bm\mu_{i}'$ and covariance $\bm\Sigma_{i}'$.
To render the color $C$ and depth $D$ of the selected pixel $x$ on the image plane, we blend $N$ ordered (by depth) 2D Gaussians overlapping the pixel:
\begin{align}
    C = \sum_{i \in N} \bm{c}_{i} \alpha_{i} \prod_{j=1}^{i-1}(1 - \alpha_{j}),
    \label{eq:rasterization}
\end{align}
\begin{align}
    D = \sum_{i \in N} z_i \alpha_{i} \prod_{j=1}^{i-1}(1 - \alpha_{j}),
    \label{eq:depth_rasterization}
\end{align}
with
\begin{equation}
\alpha_{i} = o_{i} \cdot \exp{[-\frac{1}{2}(\bm{x} - \bm{\mu'}) (\bm{\Sigma}')^{-1} (\bm{x} - \bm{\mu'})]}.
\end{equation}
Though \cref{eq:rasterization} is the same as the volume rendering used in the classical NeRF~\cite{mildenhall2021nerf},
rasterization-based rendering is much more efficient.
Because rasterization allows us to project and render objects in the space directly instead of expensive sampling along the ray used in a full volumetric representation.

\section{Methods}
\Cref{fig:enter-label} shows an overview of our method.
Given RGBD frames and estimated camera poses, we update the map by comparing the rendered images and the ground truth to identify unobserved regions and areas requiring refinement.
Regularization terms are incorporated into the optimization process to mitigate the issue of forgetting during mapping (\cref{sec:mapping}).
We track the camera pose in the Gaussian map by minimizing color and depth re-rendering loss~(\cref{sec:tracking}).

\subsection{Mapping} \label{sec:mapping}
\noindent \textbf{Gaussian Densification.}
Differently from initializing the Gaussian parameters with all images as in the original 3D Gaussian Splatting~\cite{kerbl20233dgaussian}, we process sequential RGBD frame inputs.
For the latest RGBD input frame with an estimated pose (\cref{sec:tracking}), we need to add new Gaussians to the map, either to \emph{complete unobserved regions} or to \emph{improve the rendering quality of previously mapped regions}.

First, for the unobserved regions, we need to add new Gaussians to fill holes.
We judge holes by rendering opacity.
That is, when receiving  a new frame, we render the \emph{opacity image} at the estimated pose by
\begin{align}
    O = \sum_{i \in N} \alpha_{i} \prod_{j=1}^{i-1}(1 - \alpha_{j}).
    \label{eq:occupancy_rasterization}
\end{align}
In the opacity image, if the rendered value is below some predefined opacity threshold $\tau_\mathrm{opa}$ (i.e., $O(x,y) < \tau_{\mathrm{opa}}$), we need to densify the corresponding regions to fill holes.

Second, due to e.g. occlusion and illumination changes, the rendered image is often view-dependent. In addition, the optimized Gaussian parameters may get trapped into local minima to fit certain frames.
To account for such view-dependent errors, we propose a densification step guided by \emph{rendering error}.
When giving the estimated pose, we render the color image $\hat{C}$ and the depth image $\hat{D}$ at this pose; the current reference color image and the depth image are $C$ and $D$.
We add new Gaussians when the error between the rendered image and the input image is large, where the color error and depth error are
$E_{\mathrm{color}}(x,y) = | \hat{C}(x,y) - C(x,y) |$ and $E_{\mathrm{depth}}(x,y) = |\hat{D}(x,y) - D(x,y) |$ respectively.
We will densify those regions where $E_{\mathrm{color}}(x,y) > \tau_{\mathrm{color}}$ or $\frac{E_{\mathrm{depth}}(x,y)}{D(x,y)} > \tau_{\mathrm{depth}}$,
where $\tau_{\mathrm{color}}$ and $\tau_{\mathrm{depth}}$ are predefined thresholds---we divide by $D(x,y)$ when comparing to $\tau_\text{depth}$ to compensate for varying scales.

In summary, we densify the regions where $\big(O(x,y) < \tau_{\mathrm{opa}} \big)\ | \ \big(E_{\mathrm{color}}(x,y) > \tau_{\mathrm{color}}\big) \ | \ \big(\frac{E_{\mathrm{depth}}(x,y)}{D(x,y)} > \tau_{\mathrm{depth}}\big)$.
With depth images, we project corresponding depths directly.
The experiment reported in \cref{tab:abliation_study} shows that our densification strategy can significantly improve map quality.

\hfill

\begin{figure}
    \centering
    \includegraphics[width=0.7\linewidth]{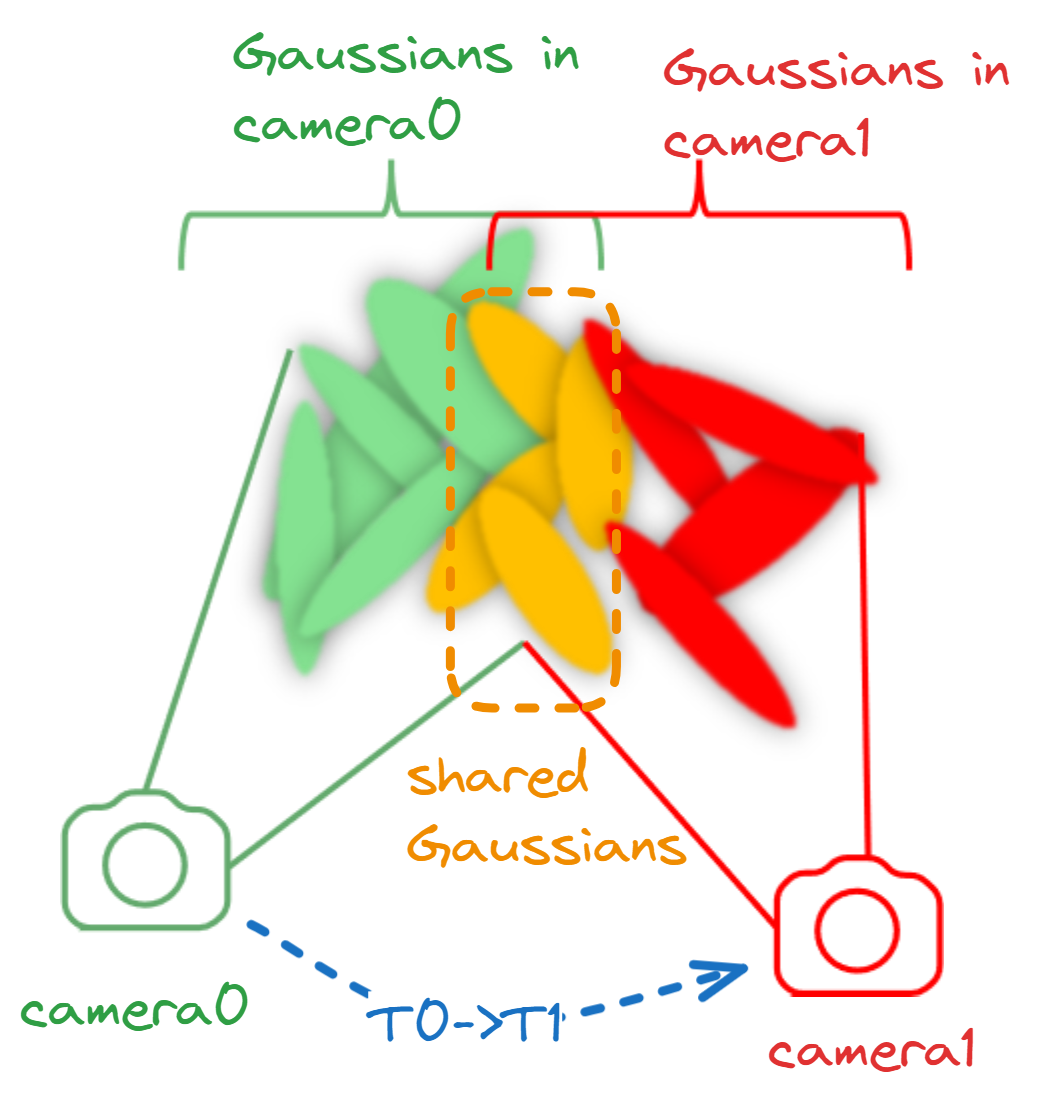}
    \caption{\textbf{Illustration of the \emph{forgetting} problem} in the context of continual mapping based on Gaussians.
    The Gaussians colored by {\color{Goldenrod!80!black}{yellow}} are shared by {\color{Green}{camera0}} and {\color{red}{camera1}}.
    However, these Gaussians tend to be optimized to overfit the latest frame {\color{red}{camera1}}, resulting in drop of reconstruction quality for previous frames.
    }
    \label{fig:forgetting-demonstration}
\end{figure}

\noindent \textbf{Continuous Mapping.}
We optimize Gaussian parameters by minimizing the loss $\ell_{1}$ between the rendered color and depth image and the input frames. In addition, following the original implementation of \textcite{kerbl20233dgaussian}, we add an extra $\mathrm{SSIM}$ term.
\begin{align}
    \mathcal{L}_{\mathrm{mapping}} = & \lambda_{\mathrm{color}} | \hat{C} - C | +  \nonumber  \lambda_{\mathrm{depth}} | \hat{D} - D| \nonumber \\
    & + \lambda_{\mathrm{SSIM}}\text{SSIM}(\hat{C},C)
    \label{eqa:mapping}
\end{align}
As a continuous learning problem, the incremental mapping in our work also suffers from the \emph{forgetting} problem~\cite{mccloskey1989catastrophic}: the same Gaussian might be ``seen'' in multiple camera poses, and the Gaussian parameters tend to be optimized (overfitted) to fit the latest frame, resulting in a performance drop for previous frames (see \cref{fig:forgetting-demonstration}).
To alleviate the overfitting problem, prior work~\cite{zhu2022nice,johari2023eslam,wang2023co} maintains a keyframe buffer from which previous keyframes are selected (based on criteria such as overlap ratio with the current frame) and the selected keyframes are then optimized together with the current frame.
In the original 3D Gaussian Splatting, all frames are queued and picked randomly for many iterations to optimize Gaussian parameters.

We can, of course, pick many previous frames whenever we optimize the current frame; however, optimizing with multiple frames leads to longer processing times, which is not suitable for online continual mapping.
Instead, to prevent Gaussian parameters from forgetting, we add an extra regularization term $\mathcal{L}_{\mathrm{reg}}$ when mapping:
\begin{align}
    \mathcal{L}_{\text{reg}} = \sum_{i \in G} \Omega_{i}^{\bm{s}} |\bm{s}_{i}^{t} - \bm{s}_{i}^{*}| + \Omega_{i}^{\bm{c}} |\bm{r}_{i}^{t} - \bm{r}_{i}^{*}| + \Omega_{i}^{d} |z_{i}^{t} - z_{i}^{*}|,
    \label{eq:regularization}
\end{align}
where $G$ refers to the set of Gaussians involved in the optimization for the current frame; $\bm{s}_{i}^{t}, \bm{r}_{i}^{t}, z_{i}^{t}$ are scales, colors and depth to be optimized at the step $t$; $\bm{s}_{i}^{*}, \bm{r}_{i}^{*}, z_{i}^{*}$ are current parameter values, that is, values achieved after step $t-1$.
$\Omega$'s are importance weights for different parameters.
In this way, we only need to maintain a small keyframe buffer but achieve better reconstruction quality.

To calculate importance weights, for each Gaussian represented by \cref{eq:gaussian_param}, we additionally add the following parameters:
\begin{align}
    \Delta_{i} \mathrel{\mathop:}= \{N_i^{\mathrm{seen}}, \Sigma_{i}^{\bm{s}}, \Sigma_{i}^{\bm{c}}, \Sigma_{i}^{\mathrm{d}} \}
\end{align}
where $N_{i}^{\mathrm{seen}}$ records the ``seen times'' of each Gaussian; $\Sigma$ are the sum of gradients of $\mathcal{L}_{\mathrm{mapping}}$ with respect to parameters, representing the sensitivity to changes.
After training each frame, we update the importance weights by averaging the gradients of mapping loss with respect to the parameters.
Taking the $i$th scale parameter $\textbf{s}_i$ as an example, after training with $N_{i}^{\mathrm{seen}}$ frames, its importance weight $\Omega_i^{\bm{s}}$ is:
\begin{align}
    \Omega_i^{\bm{s}} = \frac{\Sigma_{i}^{\bm{s}}}{N_{i}^{\mathrm{seen}}}
    = \frac{1}{N_{i}^{\mathrm{seen}}} \sum_{j}^{N_{i}^{\mathrm{seen}}} (\frac{\partial \mathcal{L}_{\mathrm{mapping}}}{\partial \bm{s}_{i}})_{j}
\end{align}
The final loss function used in the mapping is 
\begin{align}
    \mathcal{L}_{\mathrm{{mapping}}}' = \mathcal{L}_{\mathrm{mapping}} + \mathcal{L}_{\mathrm{reg}}.
\end{align}

\subsection{Tracking} \label{sec:tracking}
We conduct camera tracking in a \emph{render-and-compare} way similar to that of \emph{iNeRF}~\cite{yen2021inerf} for every input frame.
The camera pose is parameterized by $T = (t,q)$, including the 3D translation $t$ and the rotation quaternion $q$.
When tracking the new frame, we initialize the camera pose assuming constant camera velocity.
Since the color image can be affected by light, we decouple lightness from the image during tracking, similar to \cite{wang2020self6d}.
Specifically, let $\rho$ denote the operation of converting the image from the RGB space to the LAB space, and we discard the light channel~(L channel) when comparing the two images.
\begin{align}
    \mathcal{L}_{\mathrm{tracking}} = \lambda_{\mathrm{color}}|\rho(\hat{C}) - \rho(C)| + \lambda_{\mathrm{depth}} |\hat{D} - D| \label{eqa:tracking}
\end{align}

\section{Experiments} \label{sec:experiment}
In this section, we demonstrate through quantitative and qualitative experiments that our method achieves better reconstruction results than the current state-of-the-art neural implicit SLAM methods and concurrent work using Gaussian splatting.
Furthermore, the ablation studies of \cref{sec:ablation} show that our Gaussian densification strategy and regularization terms can effectively improve reconstruction quality.

\subsection{Experiment Setup} \label{sec:experiment_setup}
\noindent \textbf{Evaluation Metrics:}
We use ATE (Absolute Trajectory Error [cm]) to measure the tracking performance, which aligns the estimated trajectory and the ground truth trajectory and then computes the root mean square error.
To assess the quality of color image rendering, we employ peak signal-to-noise ratio (PSNR [dB]), structural similarity index (SSIM), and Learned Perceptual Image Patch Similarity (LPIPS). PSNR evaluates pixel-wise RGB discrepancies, SSIM gauges structural resemblance, while LPIPS quantifies perceptual image patch similarity through neural network-based analysis.
Since one can construct the mesh via \emph{TSDF-Fusion}~\cite{curless1996volumetric} with rendered depth along the estimated trajectory, we use metric Depth L1 [cm] to evaluate the geometric reconstruction as in \cite{sandstrom2023point}.
All metrics are calculated by rendering full-resolution images along the estimated trajectory after reading all frames.

\noindent \textbf{Datasets:}
We conduct our experiments on two datasets: \texttt{Replica}~\cite{straub2019replica} (a synthetic dataset)  and \texttt{TUM-RGBD}~\cite{sturm2012benchmark} (a real-world dataset); the synthetic dataset provides noiseless RGB and depth images while real-world datasets provide noisy inputs.

\noindent \textbf{Baselines:}
We evaluate our results against both state-of-the-art SLAM methods using neural implicit representation ---\emph{NICE-SLAM}~\cite{zhu2022nice}, \emph{ESLAM}~\cite{johari2023eslam}, \emph{Point-SLAM}~\cite{sandstrom2023point}---and 3D Gaussian Splatting---\emph{SplaTAM}~\cite{keetha2023splatam} and \emph{GS-SLAM}~\cite{yan2023gs}.

\noindent \textbf{Implementation Details:}
We test all the methods on one V100 GPU with 24 GB memory. The hyper-parameters used in our method are as follows: $\tau_\text{color}=0.6$, $\tau_{\text{depth}}=0.1$, $\lambda_{\text{color}}=0.5$, $\lambda_{\text{depth}}=1.0$, $\lambda_{\text{SSIM}}=0.2$.

\subsection{Reconstruction Performance}
\noindent \textbf{Quantitative Results.}
\begin{table*}
\centering
\caption{\textbf{The SLAM performance on the $\texttt{Replica}$~\cite{straub2019replica} dataset}.
The best results are highlighted by \colorbox{colorFst}{\bf first}, \colorbox{colorSnd}{second}, and \colorbox{colorTrd}{third}.
$\uparrow$ means larger is better while $\downarrow$ means smaller is better.
Our method achieves the best results in most metrics.}
\resizebox{1.0\textwidth}{!}
{
\begin{tabular}[t]{lclcccccccccc}
\toprule
\textbf{Method} & \textbf{Primitives} & \textbf{Metric} &  $\texttt{Room0}$ & $\texttt{Room1}$ & $\texttt{Room2}$ &  $\texttt{office0}$ & $\texttt{office1}$ & $\texttt{office2}$ & $\texttt{office3}$ & $\texttt{office4}$ & \textbf{Avg.}\\
\midrule

\multirow{5}{*}{\emph{NICE-SLAM}~\cite{zhu2022nice}} & \multirow{2}{*}{Neural} & PSNR [dB] $\uparrow$ & 22.12 & 22.47&24.52&29.07&30.34&19.66&22.23&24.94&24.42\\
& \multirow{2}{*}{+} & SSIM $\uparrow$ &0.69&0.76&0.81&0.87&0.89&0.80&0.80&0.86&0.81\\
& \multirow{2}{*}{Voxels} & LPIPS $\downarrow$ &0.33&0.27&0.21&0.23&0.18&0.23&0.21&0.20&0.23\\
& & ATE RMSE [cm] $\downarrow$ & 0.97 & 1.31 & 1.07 & 0.88 & 1.00 & 1.06 & 1.10 & 1.13 & 10.6\\
& & Depth L1 [cm] $\downarrow$ & 1.81 & 1.44 & 2.04 & 1.39 & 1.76 & 8.33 &  4.99 &  2.01 & 2.97\\

\hdashline

\multirow{5}{*}{\emph{ESLAM}~\cite{johari2023eslam}} & \multirow{2}{*}{Neural} & PSNR [dB] $\uparrow$ & 25.25 & 25.31 & 28.09 & 30.33 & 27.04 & 27.99 & 29.27 & 29.15 & 27.80\\
& \multirow{2}{*}{+} & SSIM $\uparrow$ & 0.87 & 0.25 & 0.93 & 0.93 & 0.91 & 0.94 & 0.95 & 0.95 & 0.92\\
& \multirow{2}{*}{Feature Plane } & LPIPS $\downarrow$ & 0.32 & 0.30 & 0.25 & 0.21 & 0.25 & 0.24 & 0.19 & 0.21 & 0.25\\
& & ATE RMSE [cm] $\downarrow$ & 0.71 & 0.70 & 0.52 & 0.57 & 0.55 & 0.58 & 0.72 & \rd 0.63 & 0.63\\
& & Depth L1 [cm] $\downarrow$ & \rd 0.97 & 1.07 & 1.28 & 0.86 &  1.26 &  1.71 & \rd 1.43 & \rd 1.06 & \rd 1.18\\
\hdashline

\multirow{5}{*}{\emph{Point-SLAM}~\cite{sandstrom2023point}} & \multirow{2}{*}{Neural} & PSNR [dB] $\uparrow$ & \rd 32.40 & \nd 34.08 & \nd 35.50 & \rd 38.26 & 39.16 & \fs 33.99 & \fs 33.48 & \nd 33.49 & \nd 35.17\\
& \multirow{2}{*}{+} & SSIM $\uparrow$ & \rd 0.97 & \fs 0.98 & \nd 0.98 & \nd 0.98 & \fs 0.99 & \rd 0.96 & \nd 0.96 & \fs 0.98 & \nd 0.97\\
& \multirow{2}{*}{Point Cloud} & LPIPS $\downarrow$ & 0.11 & \rd 0.12 &  0.11 & 0.10 &  0.12 & 0.16 &  0.13 & \rd 0.14 & 0.12\\
& & ATE RMSE [cm] $\downarrow$ & 0.61 & \rd 0.41 &  0.37 & \nd 0.38 & 0.48 & \rd 0.54 & 0.69 & 0.72 & 0.52\\
& & Depth L1 [cm] $\downarrow$ & \nd 0.53 & \fs 0.22 & \nd 0.46 & \nd 0.30 & \nd 0.57 & \fs 0.49 & \fs 0.51 & \fs 0.46 & \fs 0.44\\

\cmidrule(lr){1-12}

\multirow{5}{*}{\emph{GS-SLAM}~\cite{yan2023gs}} & \multirow{2}{*}{} & PSNR [dB] $\uparrow$ & 31.56 & 32.86 & 32.59 & \nd 38.70 & \fs 41.17 & \rd 32.36 & \rd 32.03 & \rd 32.92 & \rd 34.27\\
& \multirow{2}{*}{Parameterized} & SSIM $\uparrow$ & 0.96 & \rd 0.97 & \rd 0.97 & \nd 0.98 & \fs 0.99 & \nd 0.97 & \fs 0.97 & \nd 0.96 & \nd 0.97\\
& \multirow{2}{*}{Gaussians} & LPIPS $\downarrow$ & \rd 0.09 & \nd 0.07 & \rd 0.09 & \nd 0.05 & \fs 0.03 & \nd 0.09 & \nd 0.11 & \nd 0.11 & \nd 0.08\\
& & ATE RMSE [cm] $\downarrow$ & \rd 0.48 &  0.53 & \rd 0.33 & 0.52 & \rd 0.41 & 0.59 & \rd 0.46 & 0.70 & \rd 0.50\\
& & Depth L1 [cm] & 1.31 & \rd 0.82 & \rd 1.26 & \rd 0.81 & \rd 0.96 & \rd 1.41 & 1.53 & 1.08 & \rd 1.16 \\

\hdashline

\multirow{5}{*}{\emph{SplaTAM}~\cite{keetha2023splatam}} & \multirow{2}{*}{} & PSNR [dB] $\uparrow$ & \nd 32.86 & \rd 33.89 & \rd 35.25 & \rd 38.26 & \rd 39.17 & 31.97 & 29.70 & 31.81 & 34.11\\
& \multirow{2}{*}{Parameterized} & SSIM $\uparrow$ & \fs 0.98 & \rd 0.97 & \nd 0.98 & \nd 0.98 & \rd 0.98 & \nd 0.97 & \rd 0.95 & 0.95 & \nd 0.97\\
& \multirow{2}{*}{Gaussians} & LPIPS $\downarrow$ & \nd 0.07 & \rd 0.10 & \nd 0.08 & \rd 0.09 & \nd 0.09 & \rd 0.10 & \rd 0.12 &  0.15 & \rd 0.10\\
& & ATE RMSE [cm] $\downarrow$ & \nd 0.31 & \nd 0.40 & \nd 0.29 & \rd 0.47 & \nd 0.27 & \nd 0.29 & \nd 0.32 & \nd 0.55 & \nd 0.36\\
& & Depth L1 [cm] &--&--&--&--&--&--&--&--&-- \\

\hdashline

\multirow{5}{*}{\emph{Ours}} & \multirow{2}{*}{} & PSNR [dB] $\uparrow$ & \fs 33.06 & \fs 35.74 & \fs 37.21 & \fs 41.12 & \nd 41.11 & \nd 33.56 & \nd 33.21 & \fs 34.48 & \fs 36.19\\
& \multirow{2}{*}{Parameterized} & SSIM $\uparrow$ &\fs 0.98 & \fs 0.98 & \fs 0.99 & \fs 0.99 & \fs 0.99 & \fs 0.98 & \fs 0.97 & \fs 0.98 & \fs 0.98\\
& \multirow{2}{*}{Gaussians} & LPIPS $\downarrow$ & \fs 0.05 & \fs 0.05 & \fs 0.04 & \fs 0.03 & \fs 0.03 & \fs 0.07 & \fs 0.08 & \fs 0.08 & \fs 0.05\\
& & ATE RMSE [cm] $\downarrow$ & \fs 0.19 & \fs 0.34 & \fs 0.16 & \fs 0.21 & \fs 0.26 & \fs 0.23 & \fs 0.21 & \fs 0.38 & \fs 0.25\\
& & Depth L1 [cm] $\downarrow$ & \fs 0.39 & \nd 0.34 & \fs 0.33 & \fs 0.29 & \fs 0.26 & \nd 0.67 & \nd 0.93 & \nd 0.97 & \nd 0.52 \\

\bottomrule

\end{tabular}
}
\label{tab:reconstruction_performance_replica}
\end{table*}%

Compared to neural RGBD SLAM and more recent Gaussian Splatting SLAM, our method shows better rendering performance across both datasets, on real-world and synthetic data, as shown in \cref{tab:reconstruction_performance_replica} and \cref{tab:tum_rgbd_rendering}.
(It should be noted that since \emph{SplaTAM} and \emph{GS-SLAM} did not report rendering results on this dataset in their paper, we did not include them in the table.)

\begin{figure*}
    \centering
{\footnotesize
\setlength{\tabcolsep}{1pt}
\renewcommand{\arraystretch}{1}
\newcommand{\sz}{0.24}
\begin{tabular}{ccccc}
\multirow{2}{*}{\rotatebox[origin=c]{90}{\texttt{room 0}}} & 
\includegraphics[valign=c,width=\sz\linewidth]{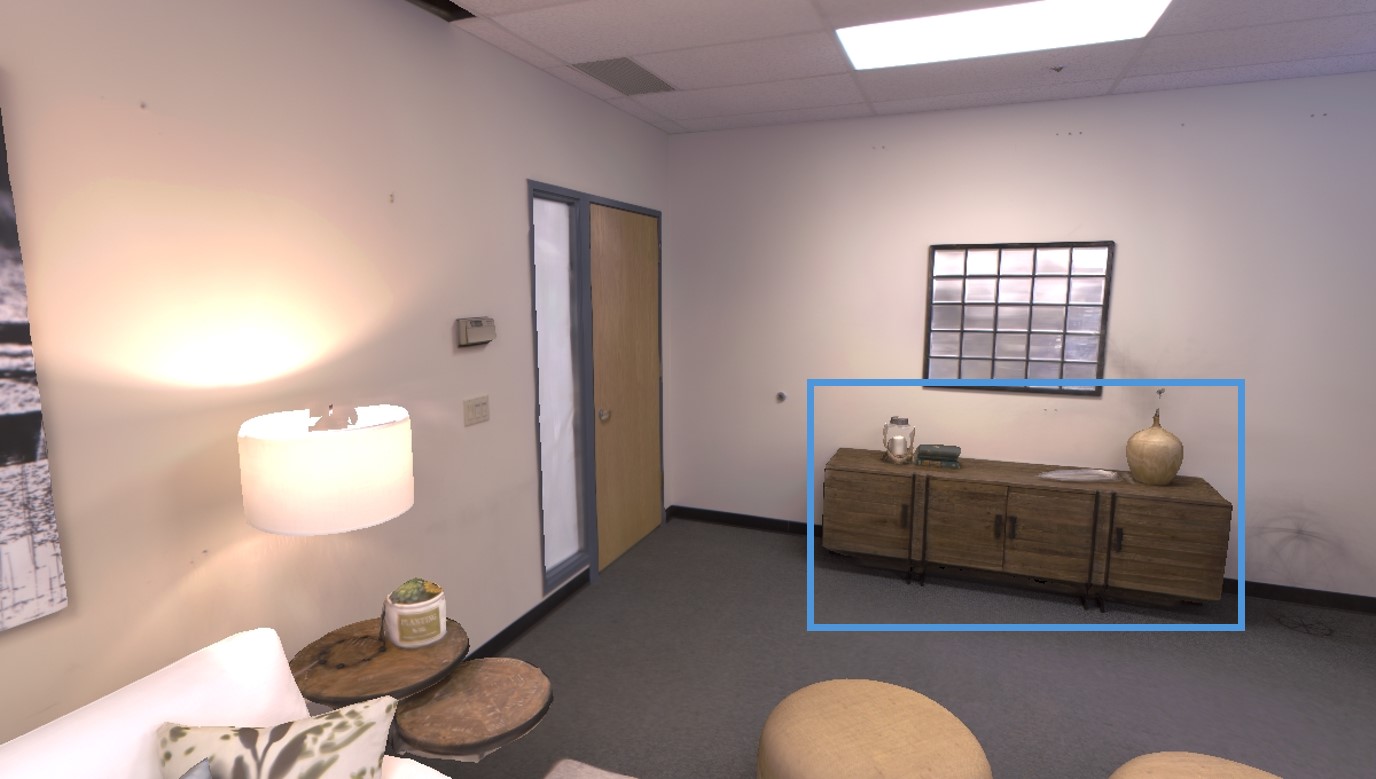} & 
\includegraphics[valign=c,width=\sz\linewidth]{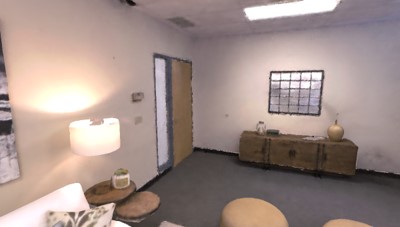} & 
\includegraphics[valign=c,width=\sz\linewidth]{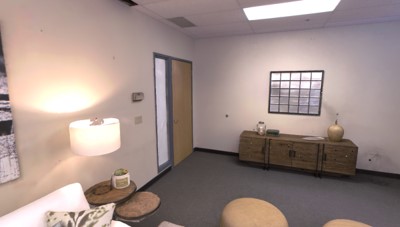} & 
\includegraphics[valign=c,width=\sz\linewidth]{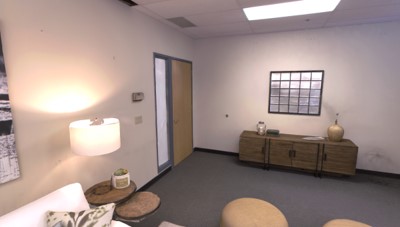}
\vspace{0.03cm}

\\
 & 
\includegraphics[valign=c,width=\sz\linewidth]{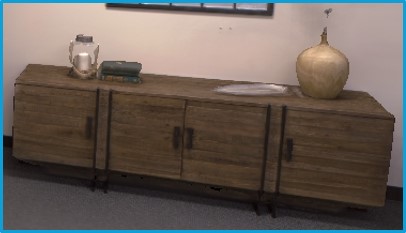} & 
\includegraphics[valign=c,width=\sz\linewidth]{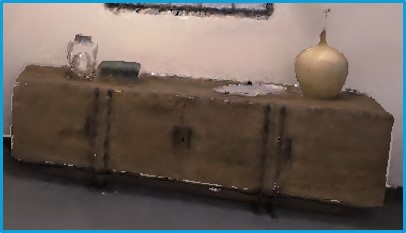} & 
\includegraphics[valign=c,width=\sz\linewidth]{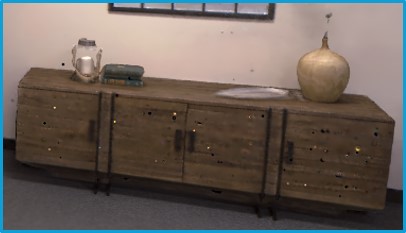} & 
\includegraphics[valign=c,width=\sz\linewidth]{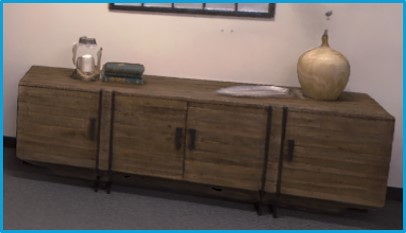}

\vspace{0.1cm}
\\

\multirow{2}{*}{\rotatebox[origin=c]{90}{\texttt{room2}}} & 
\includegraphics[valign=c,width=\sz\linewidth]{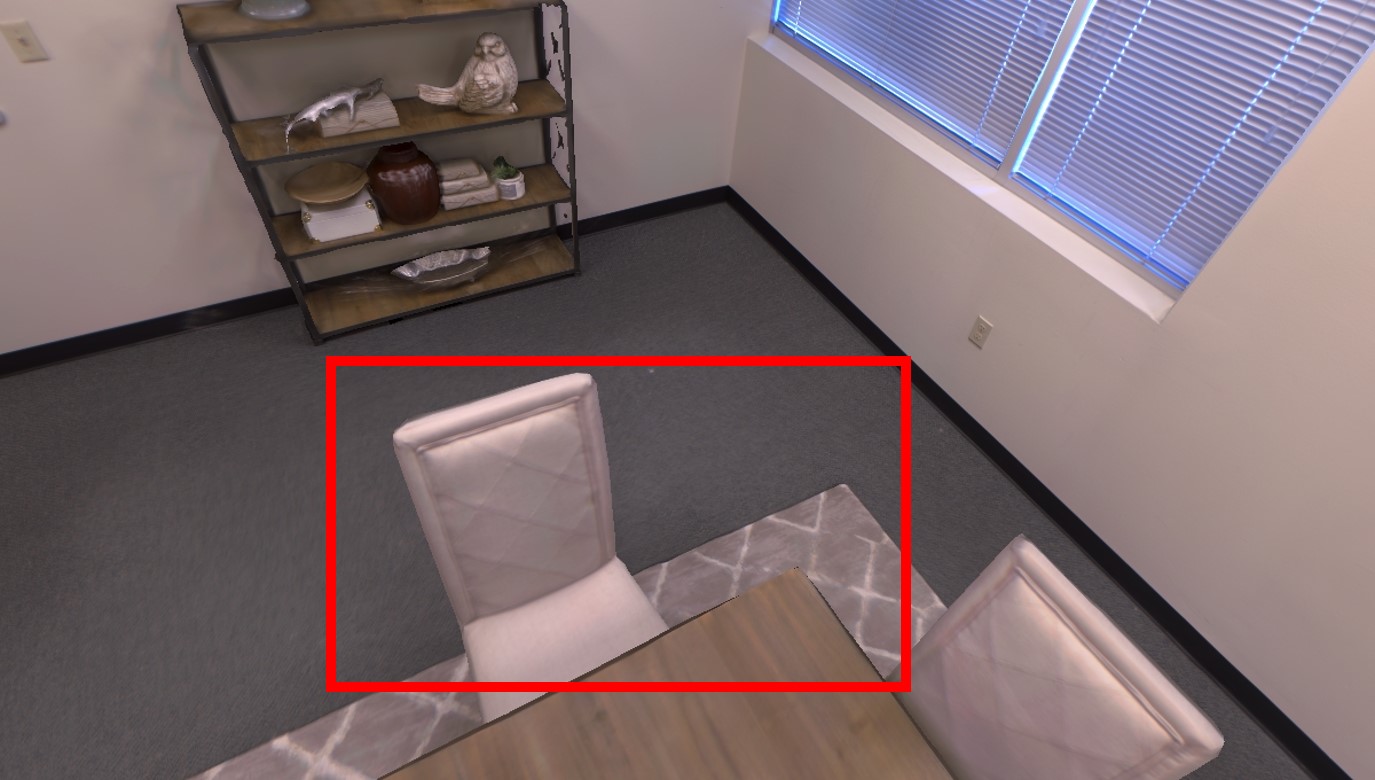} & 
\includegraphics[valign=c,width=\sz\linewidth]{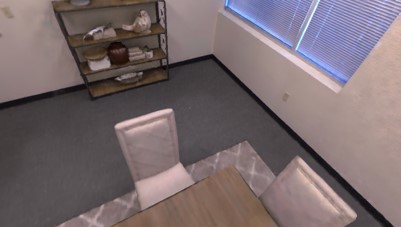} & 
\includegraphics[valign=c,width=\sz\linewidth]{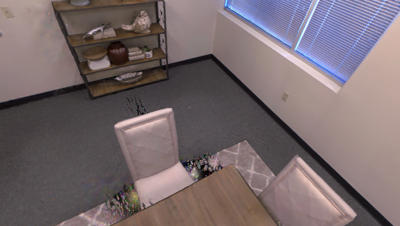} &
\includegraphics[valign=c,width=\sz\linewidth]{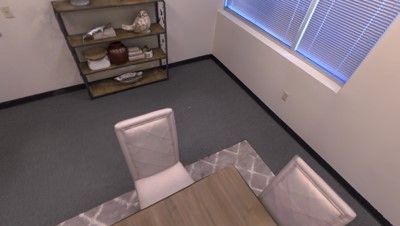} 

\vspace{0.03cm}
\\

&
\includegraphics[valign=c,width=\sz\linewidth]{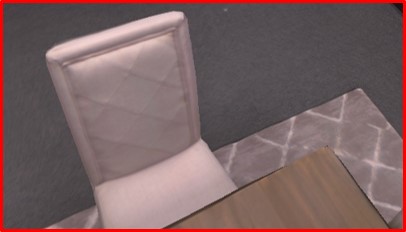} & 
\includegraphics[valign=c,width=\sz\linewidth]{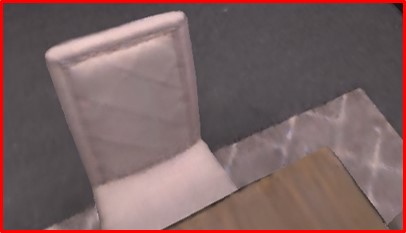} & 
\includegraphics[valign=c,width=\sz\linewidth]{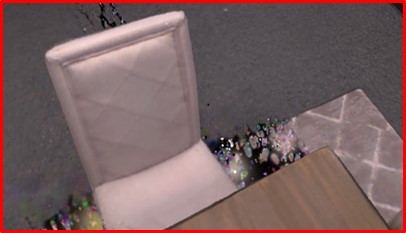} &
\includegraphics[valign=c,width=\sz\linewidth]{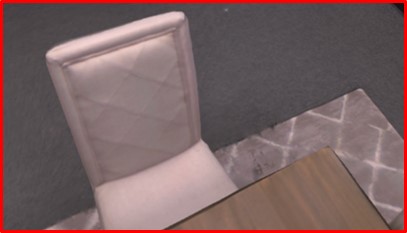}

\vspace{0.1cm}
\\
& Ground Truth & \emph{Point-SLAM}~\cite{sandstrom2023point} & \emph{SplaTAM}~\cite{keetha2023splatam} & \emph{Ours}\\
\end{tabular}
}
\caption{\textbf{The rendering results on the $\texttt{Replica}$ dataset}.
The second and forth rows are zoomed-in details of the colored squares.
Compared to \emph{Point-SLAM}~\cite{sandstrom2023point}, our method generates sharper results; compared to \emph{SplaTAM}~\cite{keetha2023splatam}, our method doen not have floaters.
}
\label{fig:rendering_results_on_replica}
\end{figure*}

Our method achieves the \colorbox{colorFst}{\bf best} results in color image rendering, but the \colorbox{colorSnd}{second} results in depth rendering in the \texttt{Replica} dataset.
To explain why our method does not consistently outperform \emph{Point-SLAM} on depth rendering, one must note that, in neural RGBD SLAM, appearance and geometry are often represented and optimized separately.
In \emph{Point-SLAM}, even though the same point cloud is used as primitives, each point has a separate color feature descriptor and a geometric feature descriptor; also, two distinct neural networks are responsible for decoding color and depth.
Hence, when optimizing the map parameters, color rendering and depth rendering do not affect each other.
However, since our method uses Gaussian Splatting, the same Gaussian parameters $\{\mathbf{\mu}, \mathbf{s}, \mathbf{r}, o\}$ are optimized for both color and depth rendering, which can impact the depth estimation.
\newline
\noindent \textbf{Qualitative Results.}
Rendered images are shown in \Cref{fig:rendering_results_on_replica}.
From the rendered results, \emph{Point-SLAM} cannot recover fine details and \emph{SplaTAM} generates floaters in the rendered image, which can be observed in the second and fourth row in \cref{fig:rendering_results_on_replica}.
On the other hand, our method generates clean images with high fidelity.

\subsection{Tracking Performance}
On the \emph{Replica} dataset, thanks to the improved reconstruction capabilities of our method shown in the previous section, we achieve the best tracking performance (see ATE RMSE in \cref{tab:reconstruction_performance_replica}).
However, on \texttt{TUM-RGBD} (\cref{tab:tum_rgbd_tracking}), our method does not always perform the best.
We hypothesise that, compared to methods using neural networks, our method is more sensitive to ``noise'', such as motion blur and exposure changes present in real data.

\begin{table}
\centering
\caption{\textbf{The rendering performance on $\texttt{TUM-RGBD}$~\cite{sturm2012benchmark} dataset}.
The best results are highlighted by \colorbox{colorFst}{\bf first}, \colorbox{colorSnd}{second}, and \colorbox{colorTrd}{third}.
}
\resizebox{1.0\linewidth}{!}
{
\begin{tabular}[t]{llcccc}
\toprule
\textbf{Method} & \textbf{Metrics} & $\texttt{fr1/desk}$ & $\texttt{fr1/desk2}$ & $\texttt{fr1/room}$ & $\texttt{fr3/off.}$ \\

\midrule

\multirow{3}{*}{\emph{NICE-SLAM}~\cite{zhu2022nice}} & PSNR [dB] $\uparrow$ & \rd 13.83 & 12.00 & \rd 11.39 & 12.89\\
& SSIM $\uparrow$ & 0.57 & 0.51 & 0.37 & \rd 0.55\\
& LPIPS $\downarrow$ & \rd 0.48 & \rd 0.52 & 0.63 & \rd 0.50\\

\hdashline

\multirow{3}{*}{\emph{ESLAM}~\cite{johari2023eslam}} & PSNR [dB] $\uparrow$ & 11.29 & \rd 12.30 & 9.06 & \rd 17.02\\
& SSIM $\uparrow$ & \nd 0.67 & \nd 0.63 & \fs 0.93 & 0.46 \\
& LPIPS & \nd 0.36 & \nd 0.42 & \fs 0.19 &  0.65 \\

\hdashline

\multirow{3}{*}{\emph{Point-SLAM}~\cite{sandstrom2023point}} & PSNR [dB] $\uparrow$ & \nd 13.87 & \nd 14.12 & \nd 14.16 & \nd 18.43\\
& SSIM $\uparrow$ & \rd 0.63 & \rd 0.59 & \rd 0.65 & \nd 0.75\\
& LPIPS $\downarrow$ & 0.54 & 0.57 & \rd 0.55 & \nd 0.45\\

\cmidrule(lr){1-6}
GS-SLAM& --& --&--& --& --\\
\hdashline
SplaTAM & --& --&--& --& --\\
\hdashline

\multirow{3}{*}{\emph{Ours}} & PSNR [dB] $\uparrow$ & \fs 22.60 & \fs 20.79 & \fs 17.53 & \fs 22.30 \\
& SSIM $\uparrow$ & \fs 0.91 & \fs 0.85 & \nd 0.72 & \fs 0.89\\
& LPIPS $\downarrow$ & \fs 0.15 & \fs 0.22 & \nd 0.32 & \fs 0.16\\

\bottomrule

\end{tabular}
}
\label{tab:tum_rgbd_rendering}
\end{table}%
\begin{table}[t]
\centering
\caption{\textbf{The tracking performance on $\texttt{TUM-RGBD}$~\cite{sturm2012benchmark} dataset}.
The best results are highlighted by \colorbox{colorFst}{\bf first}, \colorbox{colorSnd}{second}, and \colorbox{colorTrd}{third}.
The evaluation metric for the trajectory is the ATE [cm].
}
\resizebox{1.0\linewidth}{!}
{
\begin{tabular}[t]{lcccc}
\toprule
\textbf{Method} & $\texttt{fr1/desk}$ & $\texttt{fr1/desk2}$ & \texttt{fr1/room} & $\texttt{fr3/off.}$\\
\midrule

\emph{NICE-SLAM}~\cite{zhu2022nice} & \rd 4.26 & \rd 4.99 & 34.49 & \rd 3.87\\

\emph{ESLAM}~\cite{johari2023eslam} & \fs 2.47 & \fs 3.69 & \rd 29.73 & \fs 2.42\\

\emph{Point-SLAM}~\cite{sandstrom2023point} & 4.34 & \nd 4.54 &  30.92 & \nd 3.48\\
\cmidrule(lr){1-5}
\emph{GS-SLAM}~\cite{yan2023gs} & \nd 3.30 & - & - & 6.60 \\

\emph{SplaTAM}~\cite{keetha2023splatam} & 3.35 & 6.54 & \fs 11.13 & 5.16 \\

\emph{Ours} & 3.38 & 7.20 & \fs \nd 22.62 & 5.12 \\

\bottomrule

\end{tabular}
}
\label{tab:tum_rgbd_tracking}
\end{table}%
\subsection{Ablation Study}
\label{sec:ablation}
\begin{figure}[t]
    \centering
    \footnotesize
\setlength{\tabcolsep}{1pt}
\renewcommand{\arraystretch}{1}
\newcommand{\sz}{0.32}
\begin{tabular}{ccc}
\includegraphics[valign=c,width=\sz\linewidth]{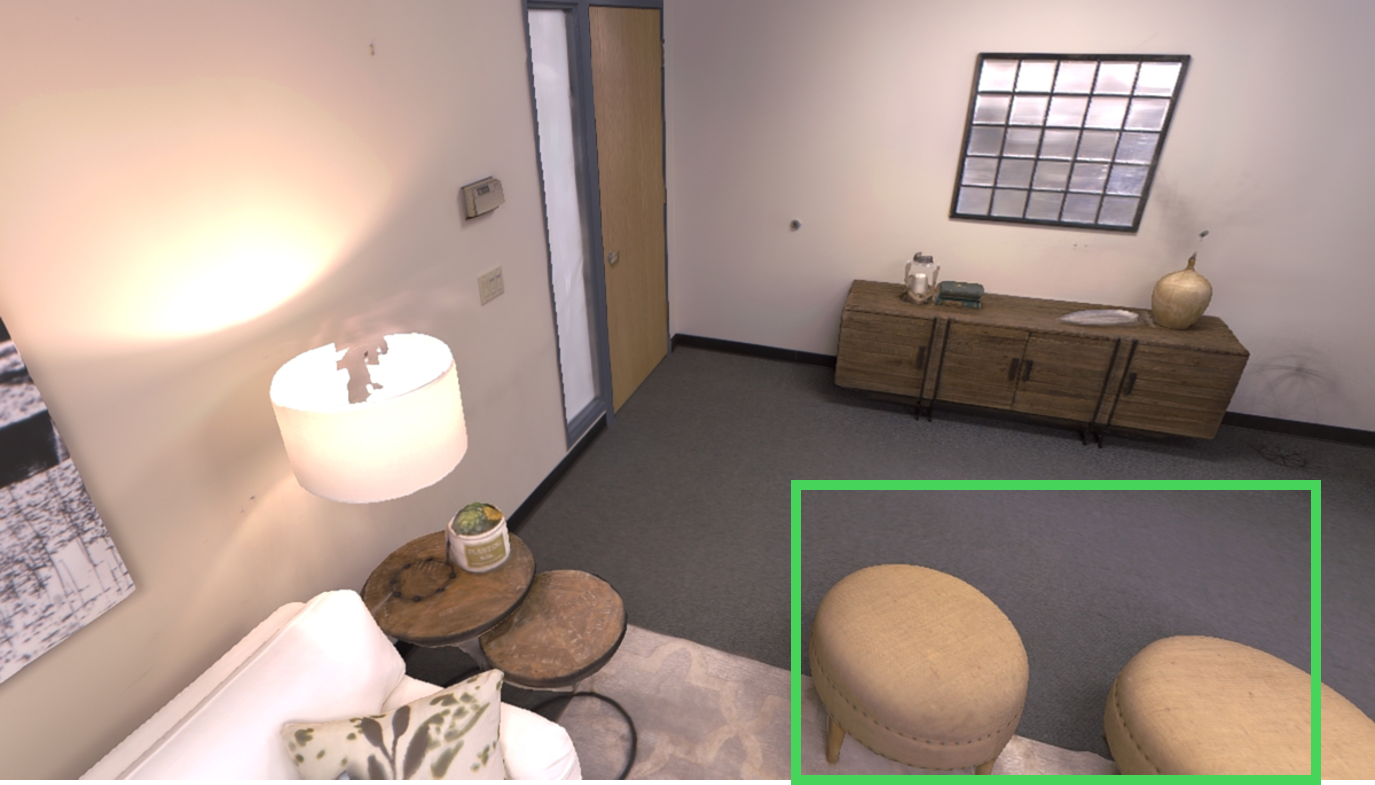} &
\includegraphics[valign=c,width=\sz\linewidth]{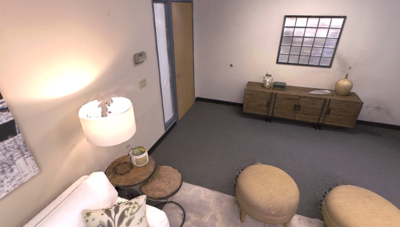} &
\includegraphics[valign=c,width=\sz\linewidth]{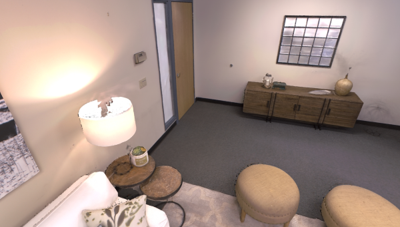}
\vspace{0.02cm}
\\

\includegraphics[valign=c,width=\sz\linewidth]{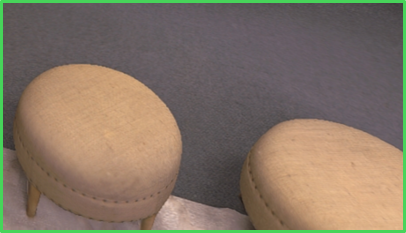} &
\includegraphics[valign=c,width=\sz\linewidth]{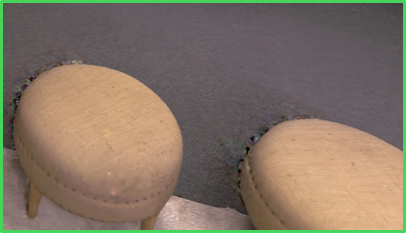} &
\includegraphics[valign=c,width=\sz\linewidth]{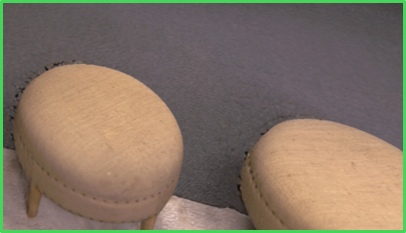}
\vspace{0.02cm}

\\

\thead{Rendered result \\ after processing \\ $\texttt{frame0}$} &
\thead{Rendered $\texttt{frame0}$ \\ after 350 frames: \\ Without Regularization}&
\thead{Rendered $\texttt{frame0}$ \\ after 350 frames: \\ With Regularization}
\end{tabular}
\caption{\textbf{The difference of without/with Regularization}.
The first column shows the rendered result after just mapping $\texttt{frame0}$.
The second column illustrates the rendered image of $\texttt{frame0}$ after processing 350 frames without regularization, while the third column showcases the same after applying regularization. 
The second rows show the zoomed-in results in the {\color{green!70!black}{green}} square, we can see that with regularization, the result can still maintain good quality, especially for edges.
}
\label{fig:abilation_study_reg}
\end{figure}

In this section, we conduct ablation studies to demonstrate the effectiveness of the regularization terms and densification.
For the regularization, we consistently render the first frame of $\texttt{Room0}$ in the \texttt{Replica} dataset at every step during the mapping.
\Cref{fig:abilation_study_reg} shows rendered results of $\texttt{frame0}$ after processing 350 frames.
In the absence of regularization, the generated image often exhibits artifacts such as floaters at the edge. Conversely, our approach ensures a cleaner output, with reduced or eliminated artifacts.

We show the overall quantitative results on the $\texttt{Room0}$ in \texttt{Replica} dataset in \cref{tab:abliation_study}.
\begin{table}
\centering
\caption{
\textbf{Quantitative results of ablation study.}
We conduct ablation studies on $\texttt{Room0}$ in \texttt{Replica} dataset.
``Regularization" means whether to add regularization terms (\cref{eq:regularization}) during mapping; ``Densification" means whether to densify regions based on color and depth rendering.
The results show our proposed methods can improve reconstruction quality and tracking accuracy.
}
\resizebox{1.0\linewidth}{!} {
\begin{tabular}[t]{cccccc}
\toprule
Regularization & {Densification} & PSNR [dB] $\uparrow$ & SSIM $\uparrow$ & LPIPS $\downarrow$ & ATE [cm] $\downarrow$\\

\midrule

{\color{red}{\XSolidBrush}} & {\color{green!80!black}{\CheckmarkBold}} & 25.96 & 0.89 & 0.17 & 0.89\\

{\color{green!80!black}{\CheckmarkBold}} & {\color{red}{\XSolidBrush}} & 25.58 & 0.85 & 0.21 & 10.02\\

{\color{green!80!black}{\CheckmarkBold}} & {\color{green!80!black}{\CheckmarkBold}} & \fs 35.74 & \fs 0.98 & \fs 0.05 & \fs 0.34\\

\bottomrule

\end{tabular}
}
\label{tab:abliation_study}
\end{table}%
With regularization terms, we can achieve better reconstruction and tracking accuracy.
Additionally, we test with and without color and depth rendering densification, and the result (see the second row in \cref{tab:abliation_study}) reveals that the color and depth rendering densification can help improve rendering quality.

\subsection{Runtime}
\begin{table}
\centering
\caption{
\textbf{Runtime results on $\texttt{Room0}$ in \texttt{Replica} dataset.}
}
\resizebox{1.0\linewidth}{!} {
\begin{tabular}[t]{lcccc}
\toprule
\textbf{Method} & \thead{Tracking [ms] \\ / Iteration} & \thead{Mapping [ms] \\ / Iteration} & \thead{Tracking [s] \\ / Frame} & \thead{Mapping [s] \\ / Frame} 

\\
\midrule

\emph{Point-SLAM}~\cite{sandstrom2023point} & \fs 31.50 & \fs 19.05 & \fs 0.63 & 7.62 \\
\emph{SplaTAM}~\cite{keetha2023splatam} & 75.25 & 91.66 & 3.01 & 5.50\\
\emph{Ours} & 45.00 & 65.57 & 1.80 & \fs 4.59\\

\bottomrule

\end{tabular}
}
\label{tab:run_time}
\end{table}%
\Cref{tab:run_time} reports the runtime results on $\texttt{Room0}$ in \emph{Replica} dataset running with a V100 GPU.
For each iteration, \emph{Point-SLAM} achieves the least time consumption because only a subset of pixels are used in the tracking and mapping (in \emph{Point-SLAM}, 1000 pixels are used in the mapping and 200 pixels are used in the tracking\footnote{\href{https://github.com/eriksandstroem/Point-SLAM}{https://github.com/eriksandstroem/Point-SLAM}}). However, such a sparse sampling needs more iterations to converge, resulting in the longest time consumption for mapping; in addition, the sparse sampling decreases the rendering quality.
\emph{SplaTAM} and our method render full-resolution images, which requires slightly more time at each iteration, but converges faster.
It should be noted that our method is faster than \emph{SplaTAM} due to implementation differences, since we render color and depth in one single rasterization process, as opposed to two rasterization processes in \emph{SplaTAM}.

\section{Summary and Future Work}

In this paper, we introduce a dense RGBD SLAM method based on the 3D Gaussian representation, which can render high-fidelity color and depth images.
Our method uses rendering-based densification, which improves the quality of reobserved parts of the environment and allows the map to be extended.
To alleviate the problem of ``forgetting'' (or overfitting) during mapping, we introduce a regularization loss in the optimization.
Experiments show that our method consistently achieves higher-quality visual reconstruction than recent baselines, both those using neural representations and those using Gaussian representations. 
An interesting observation is that splatting-based methods, including ours,  generally have lower tracking performance on real-world data (\texttt{TUM-RGBD}) compared to neural methods. However, our method still achieves the most high-fidelity visual reconstruction.

\begin{figure}[t]
    \centering
    \footnotesize
\setlength{\tabcolsep}{1pt}
\renewcommand{\arraystretch}{1}
\newcommand{\sz}{0.50}
\begin{tabular}{cc}
\includegraphics[valign=c,width=\sz\linewidth]{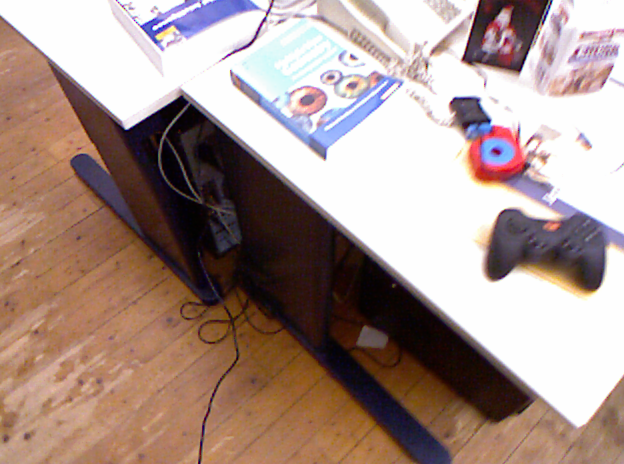} &
\includegraphics[valign=c,width=\sz\linewidth]{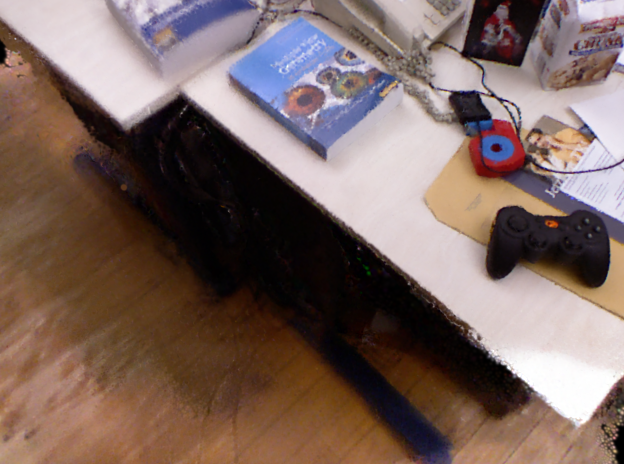} \\
Ground truth image from $\texttt{TUM}$ & \thead{The rendering image \\ PSNR: 17.16}\\
\end{tabular}
\caption{
\textbf{The ground truth image and the rendering image of $\texttt{TUM}$ dataset.}
Our method maps by accumulating all previous frames, leading to more visually pleasing image quality.
However, the ground truth image is poor in this case due to motion blur and varying exposure, which brings difficulty to tracking. Also, it negatively affects the image similarity evaluation metric: PSNR is only 17.16 in this example.
}
\label{fig:discussion_tum_varying}
\end{figure}

Despite the promising results we achieved in the experiments, our method has some limitations.
%
As shown in \cref{tab:tum_rgbd_tracking}, our method does not always achieve good tracking results in real-world datasets due to motion blur and varying exposure.
As shown in \cref{fig:discussion_tum_varying}, though the rendered image is of higher quality since the mapping accumulates all previous frames, the blurry ground truth image makes tracking difficult.
In addition, when evaluating the rendering quality, the fact that the ground truth image is blurry negatively affects the evaluation metric.

To improve tracking accuracy, it is popular to use bundle adjustment to optimize camera poses and the map simultaneously.
However, we experimentally found that simply adopting bundle adjustment as in BARF~\cite{lin2021barf} in our experiments will only make the result worse.
Encouraged by the results of \textcite{liso2024loopy}, we plan to investigate loop closure detection and optimize the trajectory with pose graph optimization.


Furthermore, future work will also investigate how to improve efficiency to make it a ``real-time" system, and, motivated by \textcite{kerr2023lerf,qin2023langsplat}, we will investigate how to introduce high-level semantic information into the mapping.


\expandafter\def\csname blx@maxbibnames\endcsname{99}%
\printbibliography
\end{document}